\documentclass[conference]{IEEEtran}
\usepackage[numbers]{natbib}
\usepackage{amsmath,amssymb,amsfonts}
\usepackage{algorithmic}
\usepackage{graphicx}
\usepackage{textcomp}
\usepackage{booktabs}
\usepackage{xcolor}
\usepackage{caption}
\usepackage[hidelinks]{hyperref}
\usepackage[hidelinks]{hyperref}
\def\BibTeX{{\rm B\kern-.05em{\sc i\kern-.025em b}\kern-.08em
    T\kern-.1667em\lower.7ex\hbox{E}\kern-.125emX}}
\begin{document}

\title{Adaptive Rollout Truncation Based on Epistemic Uncertainty for Efficient Offline World Model Training}


\author{
\IEEEauthorblockN{Nikodem Sebastian Zymla}
\IEEEauthorblockA{
\textit{TUM School of CIT}\\
Technical University of Munich\\
nikodem.zymla@tum.de
}
\and
\IEEEauthorblockN{Laurin Thiele}
\IEEEauthorblockA{
\textit{TUM School of CIT}\\
Technical University of Munich\\
laurin.thiele@tum.de
}
\and
\IEEEauthorblockN{Johannes Pitz}
\IEEEauthorblockA{
\textit{AIDX Lab}\\
Technical University of Munich\\
johannes.pitz@tum.de
}
}

\maketitle

\begin{abstract}
Accurate neural world models are central to model-based robotics, where they enable robots to predict future states from previously observed trajectories. Multi-step autoregressive training improves long-horizon prediction, but fixed rollout horizons also increase computational cost and can amplify early training errors when the model is still inaccurate. Existing training schemes typically use the same rollout length throughout optimization, independent of the model's current predictive reliability.

We propose an epistemic uncertainty-driven adaptive rollout strategy for offline world model training following an auto-curriculum training scheme. Instead of always unrolling to a fixed horizon, the model terminates autoregressive rollouts once epistemic uncertainty exceeds a threshold calibrated from a warm-up phase. We study two uncertainty estimators: a five-head ensemble with a shared recurrent backbone and Monte Carlo Dropout. A two-stage warm-up procedure stabilizes uncertainty estimates before we enable adaptive truncation.

Experiments on ANYmal-D and ANT show that ensemble-based adaptive truncation matches or improves the prediction accuracy of fixed-horizon training and the RWM-U baseline while requiring substantially fewer cumulative rollout steps. Training a world model on ANYmal-D following the presented approach reaches comparable final performance with the baselines with roughly $72$\% less rollout computation. These results indicate that epistemic uncertainty is useful not only for downstream policy regularization, but also for making world model training itself more compute-efficient.
\end{abstract}

\section{Introduction}
World models have become an increasingly important component in modern robotics. By learning the underlying system dynamics from collected trajectories, they enable robots to predict future states without interacting with the physical environment. Such predictive models can improve planning, reduce the need for expensive real-world interactions, and provide a foundation for sample-efficient model-based learning methods.

A common training strategy, as presented by \citet{li2025robotic}, for recurrent world models is autoregressive multi-step prediction. During training, the model recursively feeds its own predictions back as input to generate trajectories over multiple future time steps. This approach encourages the network to learn long-term temporal dependencies rather than only one-step transitions \cite{li2025robotic}. However, autoregressive prediction also introduces a major challenge: prediction errors accumulate over time and propagate through later rollout steps. During the early stages of training, when the model is still inaccurate, these accumulated errors can deteriorate the training signal. Long rollout horizons can thus be inefficient early in training while also requiring more computation. Most existing world model training procedures use a fixed rollout horizon throughout the complete training process. Although long rollouts become useful once the model has learned meaningful dynamics, applying the same horizon from the beginning exposes the optimizer to unreliable long-term predictions. We argue that rollout length should instead depend on model confidence.

To address this problem, we propose an epistemic uncertainty-driven adaptive rollout strategy for offline world model training. Instead of fixing the rollout horizon beforehand, the model estimates epistemic uncertainty during autoregressive prediction. Whenever the uncertainty estimate exceeds a predefined threshold, the model terminates the current rollout, and only the reliable predictions generated up to that point contribute to the loss. As training progresses and the model becomes more confident, rollouts naturally become longer without requiring a manually scheduled horizon.

We compare two epistemic uncertainty estimators within the same adaptive rollout framework: a five-head ensemble \cite{li2025uncertainty} and Monte Carlo Dropout \cite{gal2016dropout}.

We summarize the main contributions of this work as follows:
\begin{itemize}
    \item We propose an epistemic uncertainty-driven adaptive rollout termination strategy for efficient offline training of recurrent robotic world models.
    \item We compare the proposed strategy against RWM-U \cite{li2025uncertainty} and fixed-horizon rollout baselines \cite{li2025robotic} under matched compute budgets.
    \item We analyze ensemble-based and Monte Carlo Dropout-based uncertainty estimates within the same adaptive rollout framework.
    \item We show that adaptive rollout termination reduces rollout computation while maintaining or improving long-horizon prediction performance.
\end{itemize}

\section{Related Work}
Learning predictive world models has become an important research direction for improving data efficiency in robotics. \citet{li2025robotic} introduced a recurrent world model that captures robot dynamics from trajectory data through autoregressive multi-step prediction.

Building upon this work, \citet{li2025uncertainty} incorporated uncertainty estimation into the world model to improve robustness during offline model-based learning. Estimating epistemic uncertainty allows unreliable model predictions to be identified and treated differently during downstream policy training. 

As an alternative to ensembles, \citet{gal2016dropout} proposed Monte Carlo Dropout as an approximation to Bayesian inference using a single network rather than multiple independently trained models. 

Curricula over the autoregressive horizon are not themselves new. Scheduled sampling \citep{bengio2015scheduled} increases the degree of self-conditioning during training according to a hand-designed schedule. MBPO \citep{janner2019mbpo} grows its model rollout length on a fixed linear schedule on several tasks, though this controls the length of model-generated data for policy optimization rather than the autoregressive horizon used to fit the dynamics model itself. The previous work from \citet{li2025uncertainty} exploits uncertainty estimates during policy training, without adapting the world model training process itself. Our work addresses both gaps at once: we incorporate epistemic uncertainty directly into autoregressive rollout generation, using it to truncate rollouts during training rather than only as an evaluation metric or a downstream regularizer. This makes a hand-designed schedule obsolete. The horizon is set online by the model's own predictive reliability rather than by iteration count. This also reduces computation and limits the influence of unreliable long-term predictions during world model training.

\section{Method}
\subsection{World Model Architecture}
Our approach is based on the recurrent neural world model proposed by \citet{li2025uncertainty}. The architecture consists of a shared Gated Recurrent Unit (GRU) backbone followed by multilayer perceptrons (MLP). Following the work by \citet{li2025robotic} and its implementation, the model performs autoregressive multi-step predictions.

For the ensemble, five independent MLP prediction heads produce separate next-state predictions with the mean $\mu$ and variance $\sigma$. We train each head on its own bootstrap sample of the offline dataset during the warm-up, encouraging diverse predictions that capture epistemic uncertainty. For the Monte Carlo Dropout variant, the architecture remains unchanged except that dropout layers remain active during inference, and we increase the size of the MLP to match the capacity of the ensemble heads. 

\subsection{Epistemic Uncertainty Estimation}
As the world model becomes more accurate throughout training, we expect epistemic uncertainty to decrease, allowing longer and more reliable prediction horizons.

The ensemble estimates epistemic uncertainty as the variance across the means $\mu$ of five prediction heads. Large disagreement between the individual predictions indicates regions where the model is less confident. Monte Carlo Dropout estimates uncertainty by executing $10$ stochastic forward passes with dropout enabled. The variance across the resulting state predictions serves as the uncertainty estimate.

\subsection{Adaptive Rollout Truncation}
Our method dynamically determines the effective rollout length. For every batch, the model generates autoregressive predictions until either it reaches the maximum rollout length or the estimated epistemic uncertainty exceeds a dataset-specific threshold. Once the uncertainty exceeds the threshold, the model generates no further prediction steps for that rollout.

Only the successfully generated prediction steps contribute to the objective. Thus, high-uncertainty batches produce shorter rollouts, while confident batches continue until the maximum horizon.

We denote different adaptive configurations by the anchor parameter, \texttt{anchor\@2}, \texttt{anchor\@4}, \texttt{anchor\@6}, and \texttt{anchor\@8}. Here, the parameter \texttt{anchor@k} is not itself a rollout length or interval. It defines an absolute epistemic-uncertainty threshold calibrated to the uncertainty observed at the forecast horizon $k$ at the end of warm-up. During the last $10$ warm-up iterations, the model records the epistemic uncertainty at horizon $k$. The resulting threshold $T$ is the median over these 10 iterations, averaged over at least 3 independent training runs.
During training, the model truncates the rollout at the first step $i$ for which $i \geq h_{\min}$ and the mean epistemic uncertainty at step $i$ is at least $T$. We use $h_{\min}=4$ for all adaptive runs, which guarantees that gradients always flow through at least four forecast steps. Larger anchor indices produce larger thresholds and therefore later truncation, making the behavior closer to a fixed-horizon rollout.

For reproducibility across seeds, the reported runs use the frozen uncertainty
thresholds listed in Table~\ref{tab:anchor_thresholds}, rather than
re-estimating the threshold separately for every run. We calibrated the ensemble and MC Dropout thresholds after $50$ warm-up iterations.

\begin{table}[t]
    \centering
    \begin{tabular}{llcc}
        \toprule
        Dataset & Estimator & Configuration & Threshold $T$ \\
        \midrule
        ANYmal-D & Ensemble   & \texttt{anchor@2}          & 0.0808 \\
        ANYmal-D & Ensemble   & \texttt{anchor@4}          & 0.1062 \\
        ANYmal-D & Ensemble   & \texttt{anchor@6}          & 0.1216 \\
        ANYmal-D & Ensemble   & \texttt{anchor@8}          & 0.1331 \\
        ANT      & Ensemble   & \texttt{anchor@2}          & 0.1332 \\
        ANYmal-D & MC Dropout & \texttt{anchor@2}, $p=0.1$ & 0.0828 \\
        ANYmal-D & MC Dropout & \texttt{anchor@2}, $p=0.4$ & 0.1790 \\
        ANYmal-D & MC Dropout & \texttt{anchor@2}, $p=0.8$ & 0.3847 \\
        \bottomrule
    \end{tabular}
    \caption{Frozen uncertainty thresholds used for adaptive rollout truncation.}
    \label{tab:anchor_thresholds}
\end{table}

\subsection{Warm-Up Strategy}
Uncertainty estimates are difficult to interpret at the beginning of training because the model has not yet learned meaningful short-term dynamics. We therefore disable adaptive rollout truncation during the initial phase.

The default configuration first performs $50$ bootstrap single-step warm-up iterations and then $10$ full-horizon warm-up iterations (see Appendix~\ref{app:ablations} for an ablation of this choice). Restricting the rollout length to one allows the model to learn accurate local system dynamics before we introduce recursive prediction. The second warm-up phase exposes the model to long autoregressive sequences and improves prediction performance before adaptive rollout termination becomes active. Once both warm-up stages have completed, the model applies adaptive rollout termination for the remaining training process.

\section{Experimental Setup}
\subsection{Dataset}
We evaluate our method on the ANYmal-D and ANT trajectory datasets. Both datasets contain trajectories collected during PPO policy training, so the observations cover transitions from random behavior toward stronger policies. We use the data exclusively for offline world model training. We reimplemented the repository of \citet{li2025uncertainty}, since the original repository contains only the code for \textbf{online} world model training. The authors do not provide a dataset to reproduce their findings.

The ANYmal-D dataset was collected using Isaac Lab and contains $31{,}250$ segments with maximum length $200$ and mean length $192$, corresponding to approximately $6.0$ million valid transitions. Its observation and action dimensions are $45$ and $12$, respectively. It additionally provides auxiliary targets consisting of an 8-dimensional contact signal, a termination signal, and stability information. The ANT dataset was recorded with MuJoCo and contains $6{,}000$ trajectories of length $1{,}000$, corresponding to approximately $6.0$ million transitions, with observation dimension $105$ and action dimension $8$. Both datasets therefore cover a comparable number of transitions, consistent with the setup in \cite{li2025uncertainty}.

We split at the segment level, so complete trajectories go either to training or validation. We use a validation fraction of $0.1$ with a fixed split seed. For ANYmal-D this yields $28{,}125$ training segments and $3{,}125$ validation segments, or approximately $162{,}000$ training windows and $18{,}000$ validation windows. Following \citet{li2025uncertainty}, we extract training windows with $32$ history steps and $32$ forecast steps, for a total length of $64$, using stride $32$ without crossing segment boundaries. We discard segments shorter than $64$ steps.

We normalize states and actions by per-dimension z-scoring, computing the statistics from the training segments only and applying them to both training and validation data. Unless stated otherwise, we compute validation RMSE in the quantitative figures on normalized states inside the autoregressive rollout.

\subsection{Baseline}
We compare adaptive rollout training against three fixed-horizon baselines. The first is our reimplementation of the baseline RWM-U from~\citet{li2025uncertainty}. The second is our fixed-32 reimplementation, which includes the warm-up phase and performs autoregressive prediction with a constant horizon of 32 steps. The third is a fixed-8 baseline, which uses a shorter constant training horizon of $8$ steps instead after iteration $60$. Apart from the adaptive rollout mechanism and the fixed-horizon, we keep the model architecture, optimizer settings, and data identical whenever possible.

The fixed-32 experiment isolates the effect of adaptive truncation compared to the standard long-horizon training setup. The fixed-8 experiment verifies whether using a cheaper short horizon is sufficient. 

\subsection{Training Configuration}
We visualize results over different seeds with a $\pm 1\sigma$ band in the plots. For MC Dropout we test dropout rates $p=0.1$, $0.4$, and $0.8$.

\subsection{Evaluation Metrics}
First, we measure prediction accuracy using validation RMSE over the rollout horizon. Second, we evaluate per-step validation RMSE as a function of the forecast step to analyze error accumulation. Third, we approximate computational effort by the cumulative number of executed rollout steps, which measures specifically our target of adaptive truncation. Because every rollout step requires an additional recurrent forward pass, this provides a hardware-independent compute proxy. We also report measured FLOPs. The two compute metrics differ slightly because each iteration pays a fixed history-encoding cost independent of the forecast horizon.

\section{Results}

\subsection{Prediction Performance}

\begin{figure}[!t]
    \centering
    \includegraphics[width=\columnwidth]{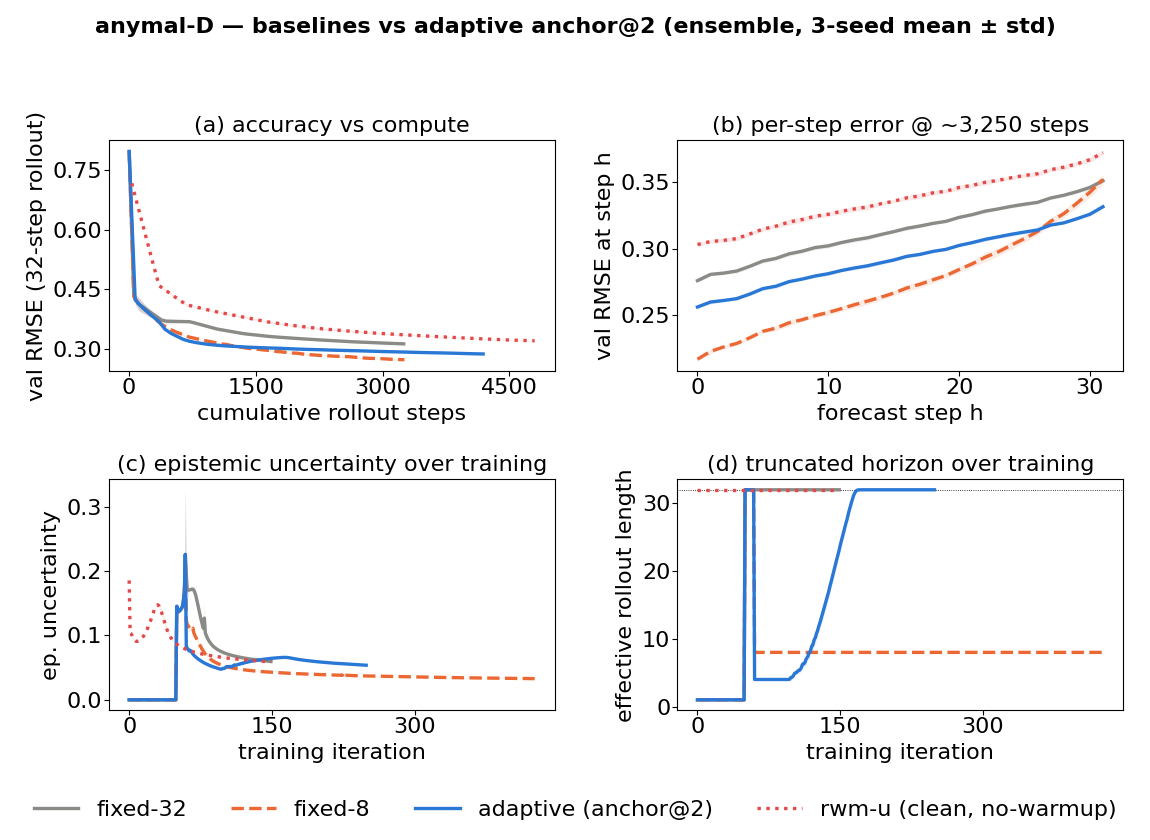}
    \caption{Adaptive \texttt{anchor@2} compared to the fixed-32 baseline, RWM-U baseline and fixed-8 baseline. Adaptive truncation and fixed-8 reach a given validation RMSE at substantially lower compute than fixed-32 and RWM-U (a), and adaptive truncation attains lower per-step error than both fixed-32 and RWM-U at equal compute (b).}
    \label{fig:fixed32_anchor2}
\end{figure}

Fig.~\ref{fig:fixed32_anchor2} compares adaptive ensemble \texttt{anchor@2} with 50 warm-up iterations, which the anchor sweep in Appendix~\ref{app:ablations} identifies as the best-performing configuration, against the fixed-32 baseline. Compared to the fixed-32 and RWM-U baseline adaptive truncation reaches the same validation RMSE with fewer cumulative rollout steps. At equal compute, it also achieves lower 32-step prediction error. The rollout statistics show the intended behavior: epistemic uncertainty decreases during training, while the number of executed rollout steps increases.


\begin{figure}[!t]
    \centering
    \includegraphics[width=0.90\columnwidth]{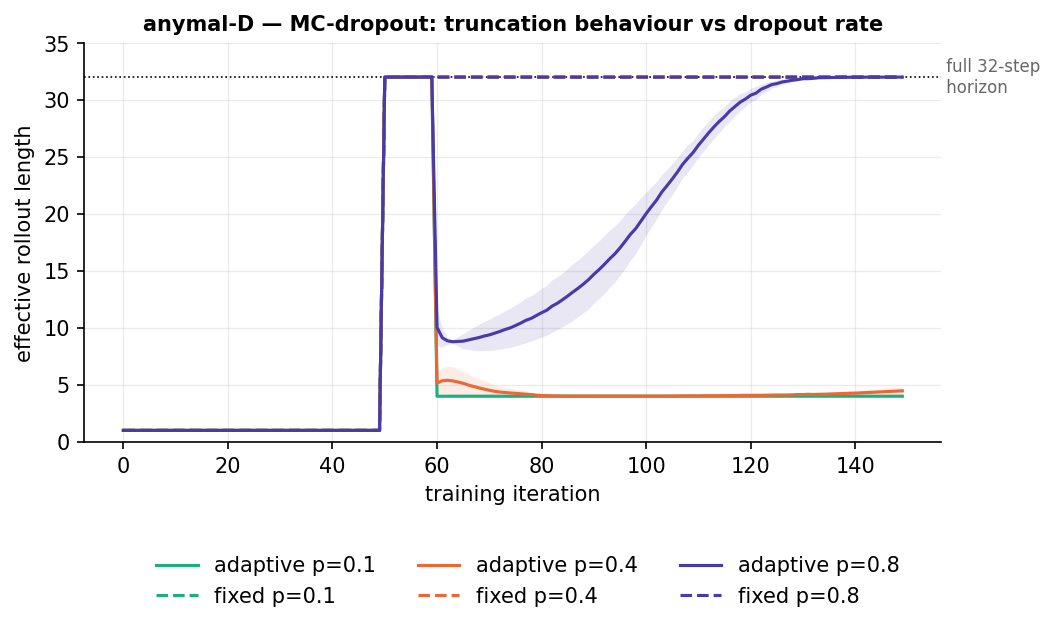}
    \caption{\textbf{Adaptive truncation using MC Dropout.}}
    \label{fig:mcdropout}
\end{figure}

Fig.~\ref{fig:fixed32_anchor2} also compares adaptive truncation with a fixed-8 baseline. 
The fixed-8 model performs well on early prediction steps and reaches a better accuracy at convergence. But its error grows more strongly beyond its training horizon due to its fixed training scheme. In contrast, adaptive truncation first focuses on both reliable short rollouts and later shifts toward accurate long-horizon predictions. Appendix~\ref{app:performance} illustrates how the compounding error develops during extended rollouts.

Table~\ref{tab:results} summarizes the final prediction accuracy and the cumulative rollout steps at convergence for all methods. It also includes a final evaluation over long forecasts. Consistent with the reduction in rollout steps, adaptive truncation also directly reduces FLOPs, see Appendix \ref{app:flops}.

Fig.~\ref{fig:mcdropout} shows the Monte Carlo Dropout experiments. 
Adaptive truncation only becomes clearly active for $p=0.8$, where the number of rollout steps taken increases. For the other configurations the minimal number of rollout steps $h_{min}=4$ is dominant. MC Dropout adds a dependency on the dropout rate $p$, thus the ensemble is preferred.

Further ablations regarding the prediction performance are described in Appendix \ref{app:performance}.

\subsection{ANT environment}
\begin{figure}[!t]
    \centering
    \includegraphics[width=0.90\columnwidth]{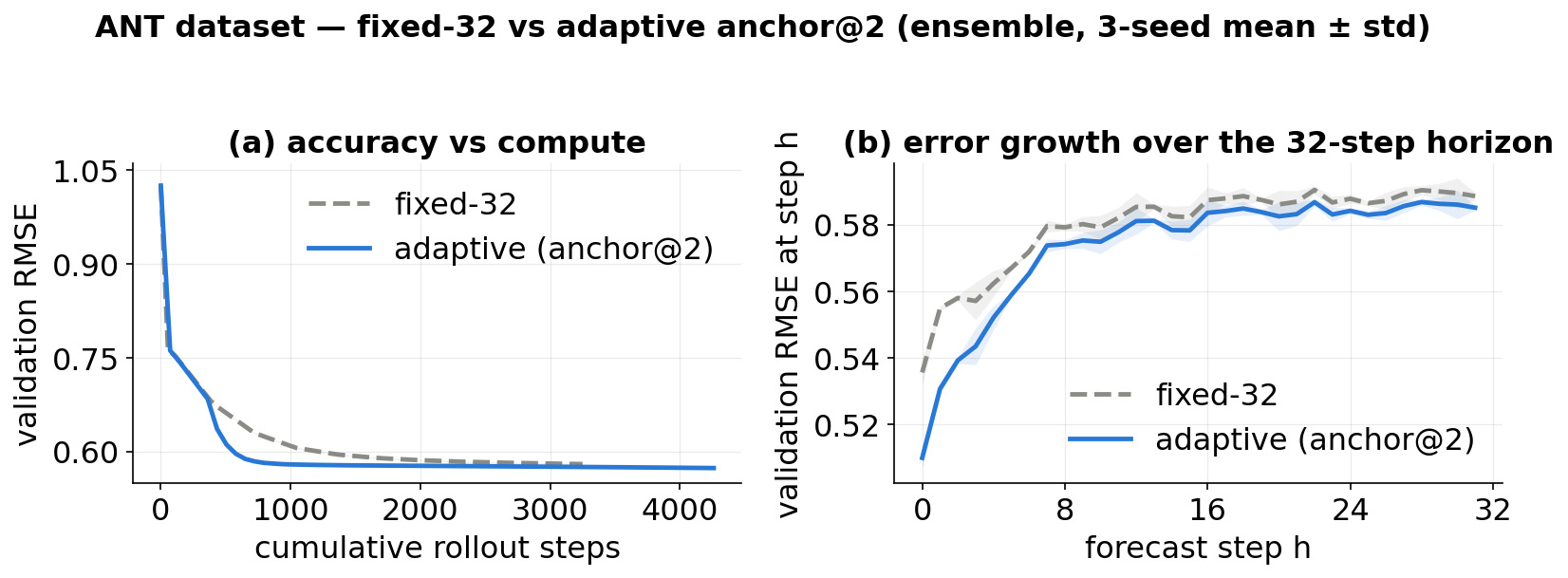}
    \caption{\textbf{Adaptive truncation trained on the ANT dataset.}}
    \label{fig:ant_ablation}
\end{figure}

We verify the method by transferring the ensemble approach to the ANT environment. We relaunch the fixed baseline runs to measure the \texttt{anchor@2} value. Fig. \ref{fig:ant_ablation} shows that adaptive truncation again converges faster during training, with better accuracy at equal compute over the 32-step prediction.

\section{Discussion}
Our method introduces a dependency on the uncertainty threshold. While we observe that the approach is relatively robust to reasonable threshold variations, overly conservative or aggressive thresholds may lead to suboptimal trade-offs between compute efficiency and long-horizon learning.




Further, \citet{li2025uncertainty} report that an offline-trained policy can outperform the data-collecting policy on real robots. Their released code provides only online world-model and policy training rather than an offline pipeline, and their pretrained model does not match our accuracy on our data (Appendix~\ref{app:external_wm}). In contrast, our contribution is compute-efficient offline world-model training via adaptive truncation. We validate its downstream use by showing, in simulation, that an epistemic penalty closes the world-model-exploitation gap (imagined-to-sim) as a function of $\lambda$. We deploy the trained \texttt{anchor@2} model in an uncertainty-penalized MOPO-PPO setup, reported in Appendix~\ref{app:mopo}.



\section{Conclusion}
In this work, we presented an uncertainty-driven adaptive rollout truncation strategy for offline training of recurrent neural world models, which adjusts the effective rollout length based on epistemic uncertainty from either ensembles or Monte Carlo Dropout under an auto-curriculum training scheme. We showed that this simple mechanism leads to a substantial reduction in computational cost while maintaining prediction accuracy comparable to fixed-length rollouts. When transitioning from short to long rollouts during training, the compounding error over long forecasts is smaller compared to short fixed-length training.

These results suggest that uncertainty-aware control of the training process itself is a promising direction for improving the efficiency of model-based reinforcement learning systems. Future work may explore alternative uncertainty measures, automatic threshold selection, and evaluation on additional robotic benchmarks such as different locomotion tasks and environments.



\begin{table}[!t]
    \centering

    \resizebox{\columnwidth}{!}{%
        \begin{tabular}{lcccc}
            \hline
            Method
            & 32-step RMSE
            & 160-step RMSE
            & Cum. rollout steps
            & $\Delta$ compute \\
            \hline
            fixed-32
            & $0.314 \pm 0.001$
            & $0.429 \pm 0.021$
            & 3{,}250
            & --- \\

            fixed-8
            & $0.274 \pm 0.002$
            & $0.458 \pm 0.006$
            & 1{,}090
            & $-66\%$ \\

            RWM-U~\cite{li2025uncertainty}
            & $0.321 \pm 0.001$
            & $0.419 \pm 0.007$
            & 4{,}800
            & $+48\%$ \\

            Adaptive \texttt{anchor@2}
            & $0.288 \pm 0.002$
            & $\mathbf{0.389 \pm 0.007}$
            & 924
            & $\mathbf{-72\%}$ \\
            \hline
        \end{tabular}%
    }
    \caption{ANYmal-D prediction accuracy and compute using an ensemble estimator (three-seed mean $\pm$ std). Compute is relative to fixed-32 accuracy ($32$ steps) and the RWM-U results are reported at convergence.
}
    \label{tab:results}
\end{table}

\bibliographystyle{IEEEtranN}
\bibliography{refs}
\clearpage
\appendices
\section{Anchor and Warm-Up Ablations}
\label{app:ablations}

\begin{figure}[h]
    \centering
    \includegraphics[width=\columnwidth]{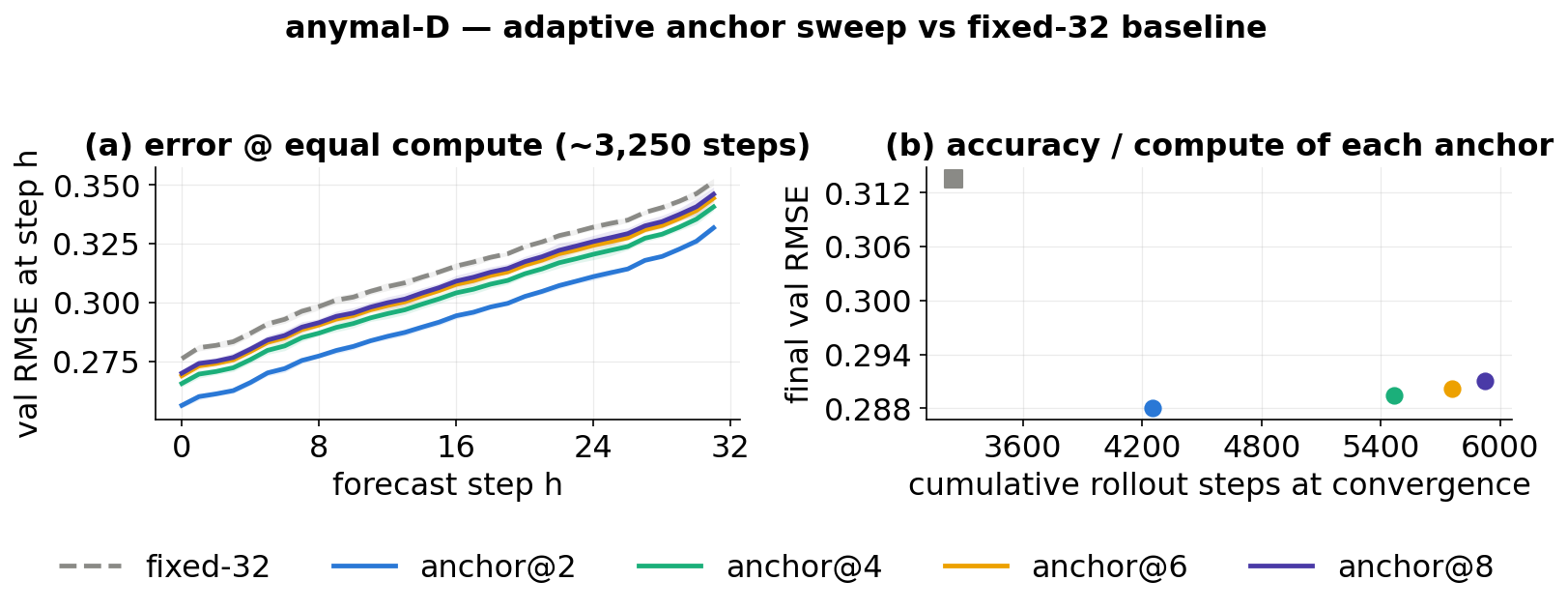}
    \caption{\textbf{Adaptive truncation for different epistemic thresholds} measured at different anchors. Comparison of model configurations under matched compute budgets (left) and at convergence (right).}
    \label{fig:anchor_sweep}
\end{figure}

Fig.~\ref{fig:anchor_sweep} compares the anchor thresholds evaluated for the ensemble listed in Table~\ref{tab:anchor_thresholds}.
\texttt{Anchor@2} applies the strictest truncation and therefore uses the fewest rollout steps, while
also reaching the lowest 32-step validation error. Larger anchor indices allow longer rollouts
earlier in training, but do not improve prediction accuracy. We therefore use \texttt{anchor@2} for all
main experiments.

\begin{figure}[h]
    \centering
    \includegraphics[width=\columnwidth]{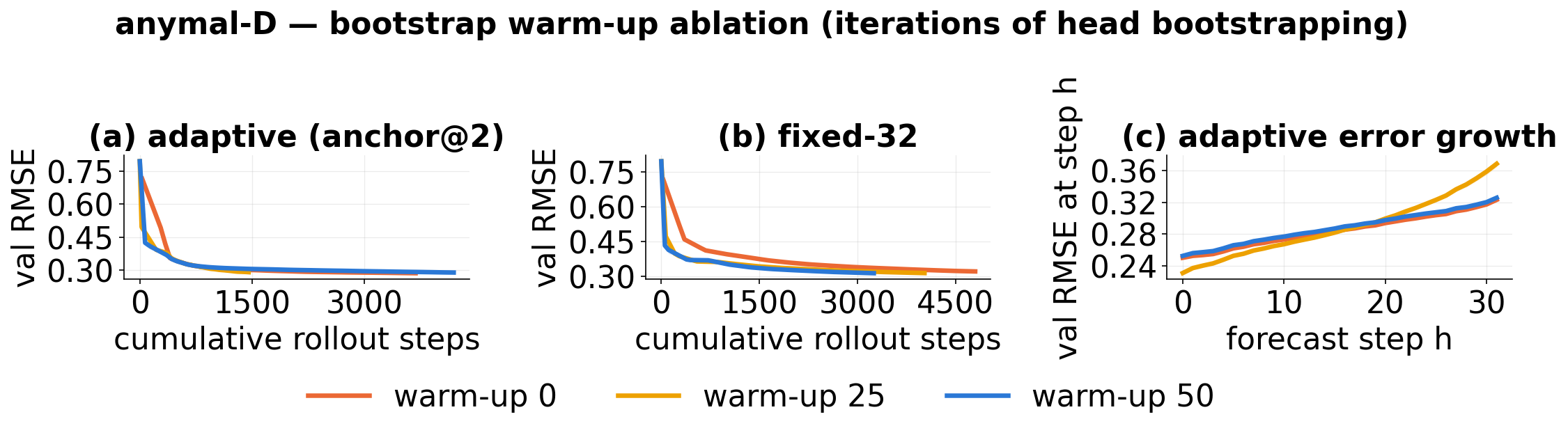}
    \caption{\textbf{Adaptive truncation for different warm-up lengths.} A warm-up phase of 50 iterations matches the results of no bootstrap warm-up at all. Training with a warm-up window of 25 steps yields the weakest results.}
    \label{fig:warmup_ablation}
\end{figure}

Fig.~\ref{fig:warmup_ablation} studies the number of bootstrap warm-up iterations. The warm-up length
has only a small effect on final accuracy, so adaptive truncation for the ensemble is robust to this choice.
We nevertheless keep 50 warm-up iterations as the default to ensure that the ensemble heads
are sufficiently developed before we enable adaptive truncation.


\section{Performance Ablation}
\label{app:performance}

\begin{figure}[h]
\centering
\includegraphics[width=\columnwidth]{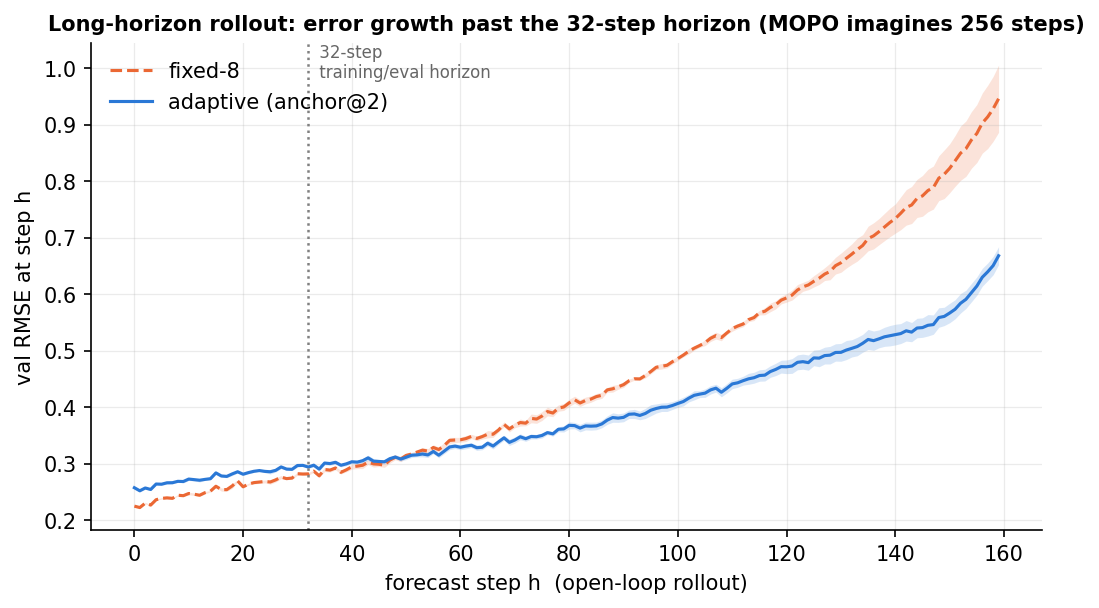}
\caption{\textbf{Long-horizon prediction error} of fixed-8 and adaptive \texttt{anchor@2}, evaluated on $3{,}000$ held-out rollouts. The fixed-8 model focused on shorter rollouts during training, its error increases dramatically on long forecasts. Our proposed auto-curriculum uncertainty scheme transitions from short to long rollouts and shows a gentle increase in the RMSE error.}
\label{fig:long_forecasts}
\end{figure}

As reported in Table~\ref{tab:results}, the prediction error of the fixed-8 baseline increases more rapidly over long forecast horizons. Fig. \ref{fig:long_forecasts} compares the two training schemes and illustrates how compounding error develops during extended rollouts. This setting is particularly relevant because downstream policy training relies on long model-generated trajectories. Since our adaptive method trains on rollouts of varying lengths, it remains more stable over long forecast horizons.

\begin{figure}[h]
\centering
\includegraphics[width=\columnwidth]{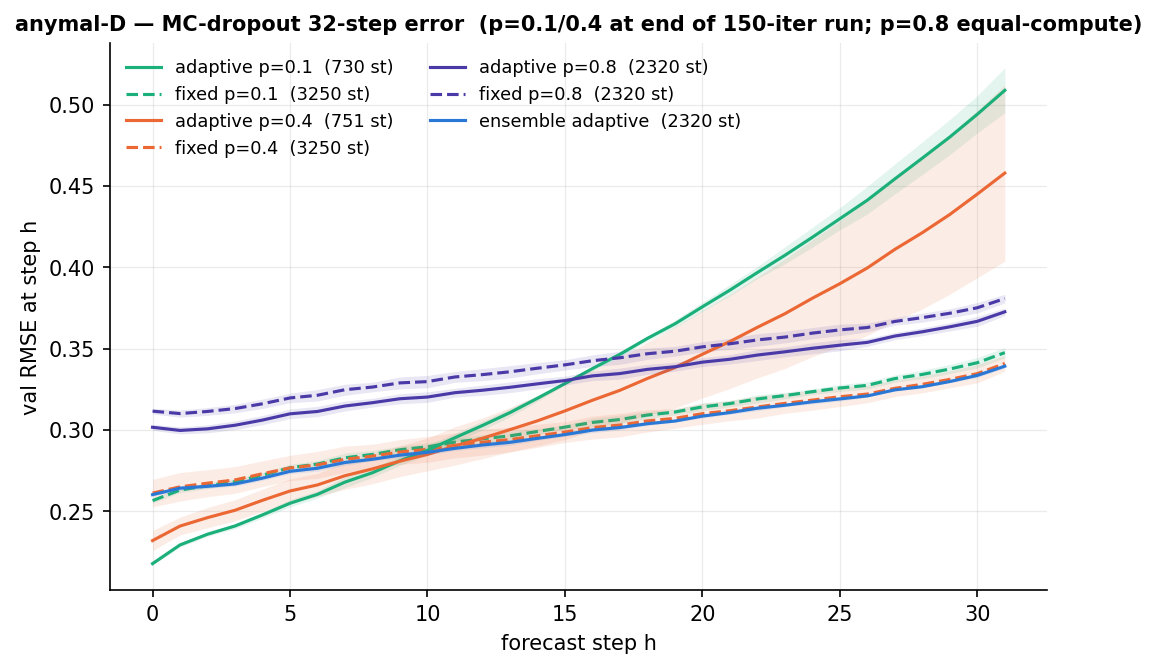}
\caption{\textbf{Validation error across the 32-step prediction horizon for different MC Dropout rates.} For $p=0.1$ and $p=0.4$, adaptive truncation does not engage. Thus, the model trains only with the minimal rollout length, and its forecast error increases dramatically. The model with $p=0.8$ is trained on both short and long rollouts, but its accuracy over the 32 forecast steps is higher than that of its fixed and adaptive ensemble counterparts.}
\label{fig:mc_dropout_val}
\end{figure}

\begin{figure}[h]
\centering
\includegraphics[width=\columnwidth]{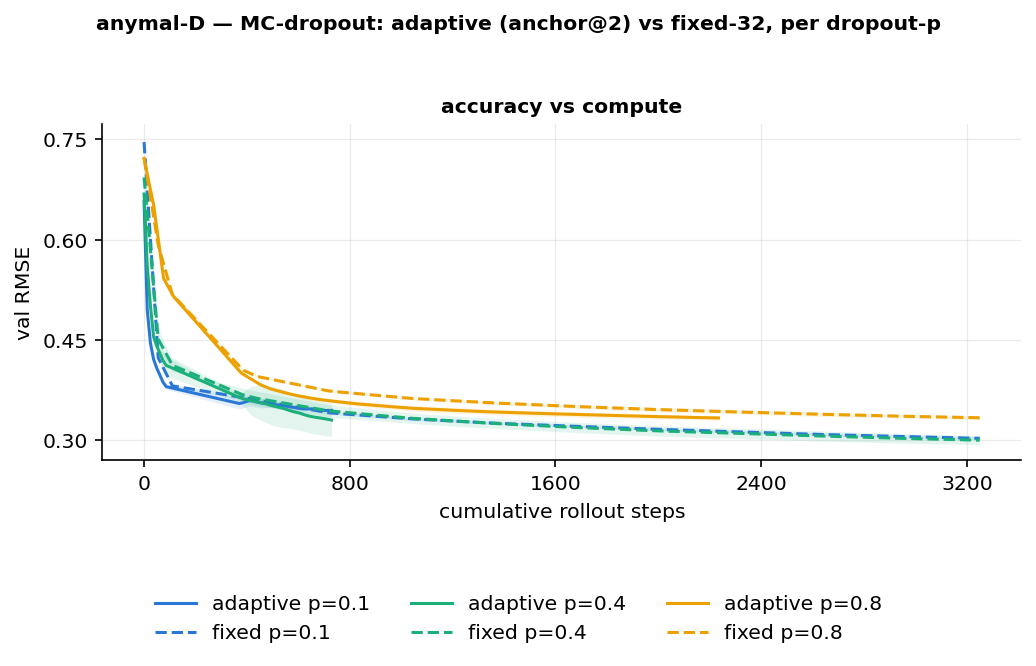}
\caption{\textbf{Adaptive truncation training for the MC Dropout estimator.} The models with $p=0.4$ and $p=0.8$ train only with the minimal rollout length. Thus, they show faster convergence and a lower RMSE error compared to the adaptive model with $p=0.8$, which is trained on the transition from short to long rollouts.}
\label{fig:mc_dropout_verlauf}
\end{figure}

For MC Dropout, adaptive truncation is activated only at the high dropout rate $p=0.8$. The dropout probability therefore introduces an additional sensitive hyperparameter for the uncertainty estimator. Fig. \ref{fig:mc_dropout_val} shows the validation error across the 32-step prediction horizon for different dropout rates. Compared with the corresponding fixed-32 baseline MC Dropout achieves only a marginal improvement. Further, compared with the adaptive \texttt{anchor@2} ensemble, MC Dropout achieves less competitive prediction performance at equal compute. Fig. \ref{fig:mc_dropout_verlauf} visualizes the training progress for the MC Dropout estimator and the final accuracy at convergence for $p=0.8$. These results provide further motivation for using an ensemble as the uncertainty estimator for adaptive rollout truncation.

\subsection{Qualitative Rollouts}
  \label{sec:qualitative}
\begin{figure}[h]
    \centering
    \includegraphics[width=\columnwidth]{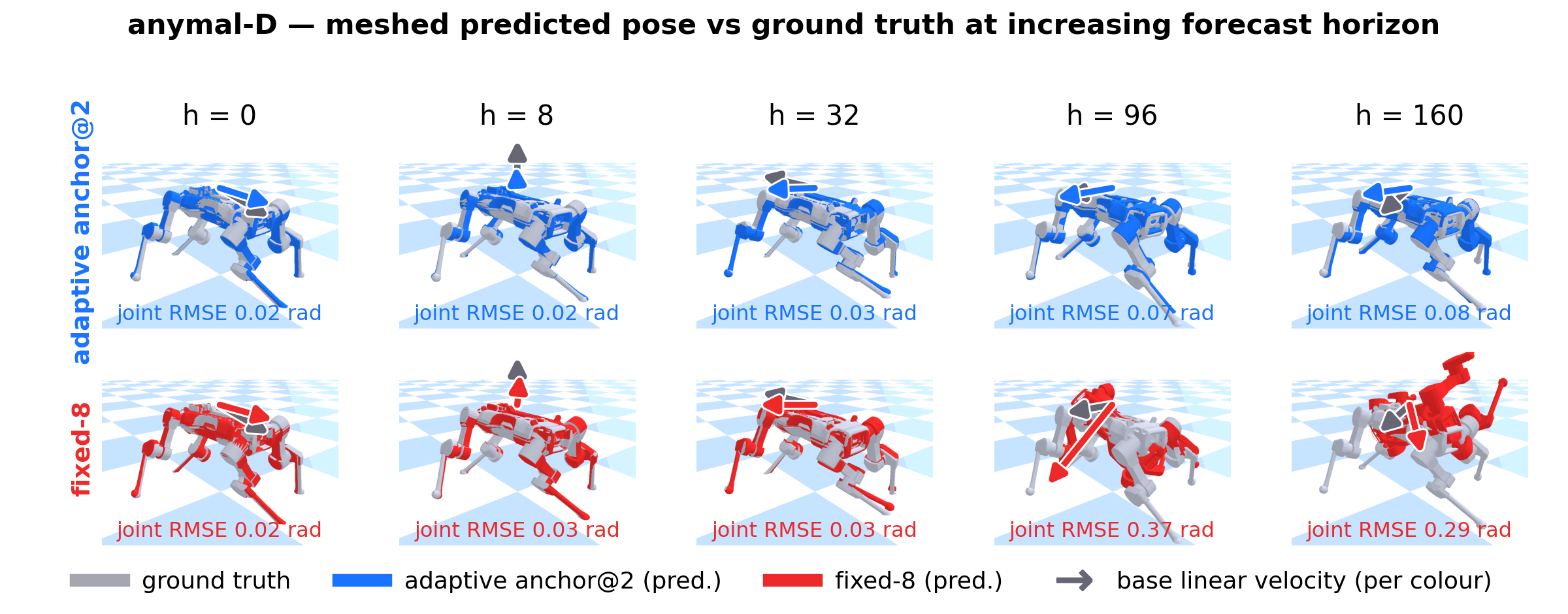}
    \caption{\textbf{Qualitative open-loop rollout as robot pose.} The ANYmal-D
      rendered at the world model's predicted joint angles at forecast horizons
      $h\in\{0,8,32,96,160\}$. Both models track the truth up to
      the $32$-step training horizon. The fixed-$8$ baseline then collapses, whereas the adaptive model remains coherent through $160$ steps.
      The value under each frame is the per-frame RMSE over the $12$ joint angles.
      Poses are anchored at a common base position, so deviation reflects joint, tilt, and velocity error.
      Rendered in PyBullet from the public ANYmal-D description.}
    \label{fig:pose_film}
  \end{figure}

  To make the open-loop error concrete, Fig.~\ref{fig:pose_film} renders the
  predicted robot state as a physical pose. Starting from a $32$-step history,
  each world model is rolled out autoregressively for up to $160$ steps (the
  MOPO imagination depth) on a held-out validation segment. We pose the
  ANYmal-D at the \emph{predicted} joint angles $q$ (state dimensions $9$--$20$)
  and recover the base \emph{roll and pitch} from the predicted projected-gravity
  vector. Yaw is not identifiable from projected gravity (a rotation about the
  gravity axis leaves the vector unchanged) and is absent from the observation,
  so it is fixed to zero. The ground-truth
  pose is overlaid as a translucent ghost, and because the absolute base position
  is likewise unobservable, both poses are anchored at a common origin. All
  visible deviation reflects joint-, tilt-, and velocity-prediction error rather
  than global translation or heading. Arrows denote the base linear velocity, in
  the same colour as each robot (grey: ground truth, blue/red: prediction).
  Up to the $32$-step training horizon both models track the ground truth closely
  (per-joint RMSE $\leq 0.03$\,rad). Beyond it, the fixed-$8$ baseline diverges catastrophically: its pose detaches from
  the ground truth and collapses by $h\!=\!96$, while the adaptive model stays
  pose-coherent through the full $160$-step rollout ($\approx0.08$\,rad). This is the qualitative counterpart of
  the long-horizon error reported in Fig.~\ref{fig:long_forecasts} and
  Table~\ref{tab:results}.

\section{MOPO-PPO Reproduction}
\label{app:mopo}

\begin{figure}[h]
    \centering
    \includegraphics[width=\columnwidth]{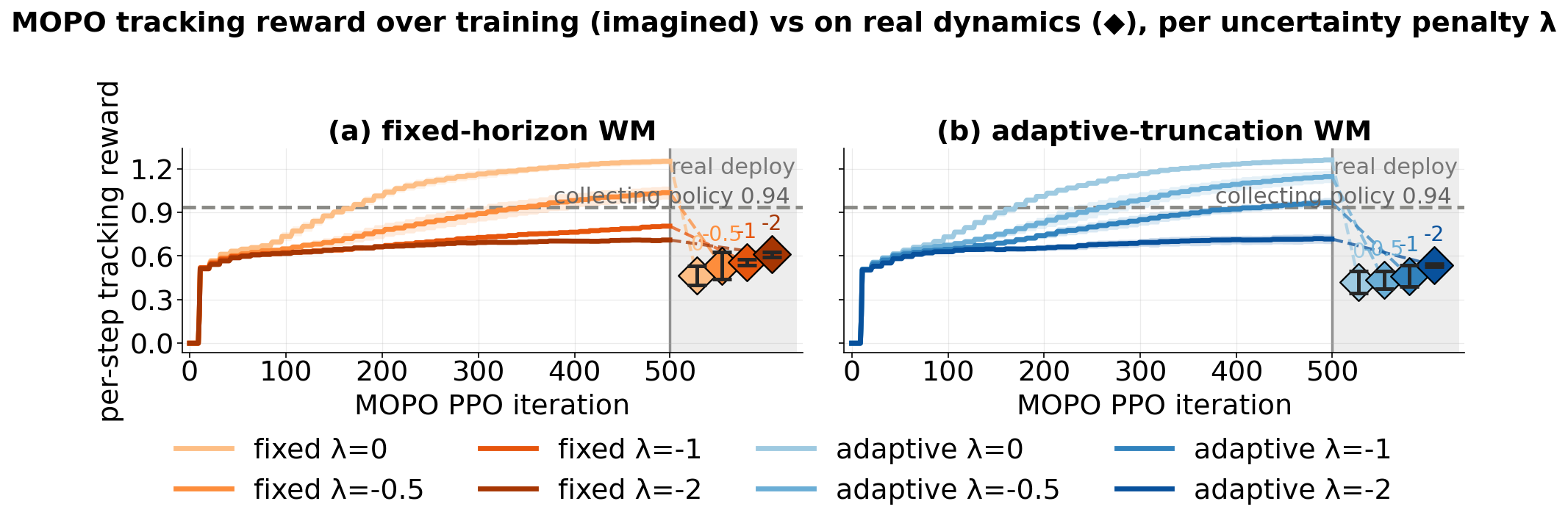}
    \caption{Deploying the trained \texttt{anchor@2} world model into MOPO-PPO training presented in \citep{li2025uncertainty}. The imagined-to-sim gap is reduced by a higher value of $\lambda$ for both the fixed ensemble (a) and adaptive model (b).}
    \label{fig:mopo_lambda}
\end{figure}

We reproduce the uncertainty-penalized MOPO-PPO setup that \citet{li2025uncertainty}
use for downstream evaluation, and deploy our trained \texttt{anchor@2} ensemble world model in it.
Fig.~\ref{fig:mopo_lambda} reports the tracking reward over MOPO-PPO iterations for the imagined
rollouts and for the corresponding evaluation on simulated dynamics, for several uncertainty
penalties $\lambda$.

Without an epistemic penalty, the policy reaches a higher imagined model reward but visits
more uncertain states. A stronger penalty reduces epistemic uncertainty but also lowers the
imagined task reward. The gap between imagined and simulated return therefore shrinks as
$\lambda$ increases, which matches the findings of \citet{li2025uncertainty}.
We do not observe a clear difference in downstream policy training between the
adaptive-truncation and the fixed-horizon world model. This is consistent with the scope of
this work, which targets world model training rather than policy optimization.

\section{Pretrained World Model}
\label{app:external_wm}
\begin{figure}[h]
    \centering
    \includegraphics[width=\columnwidth]{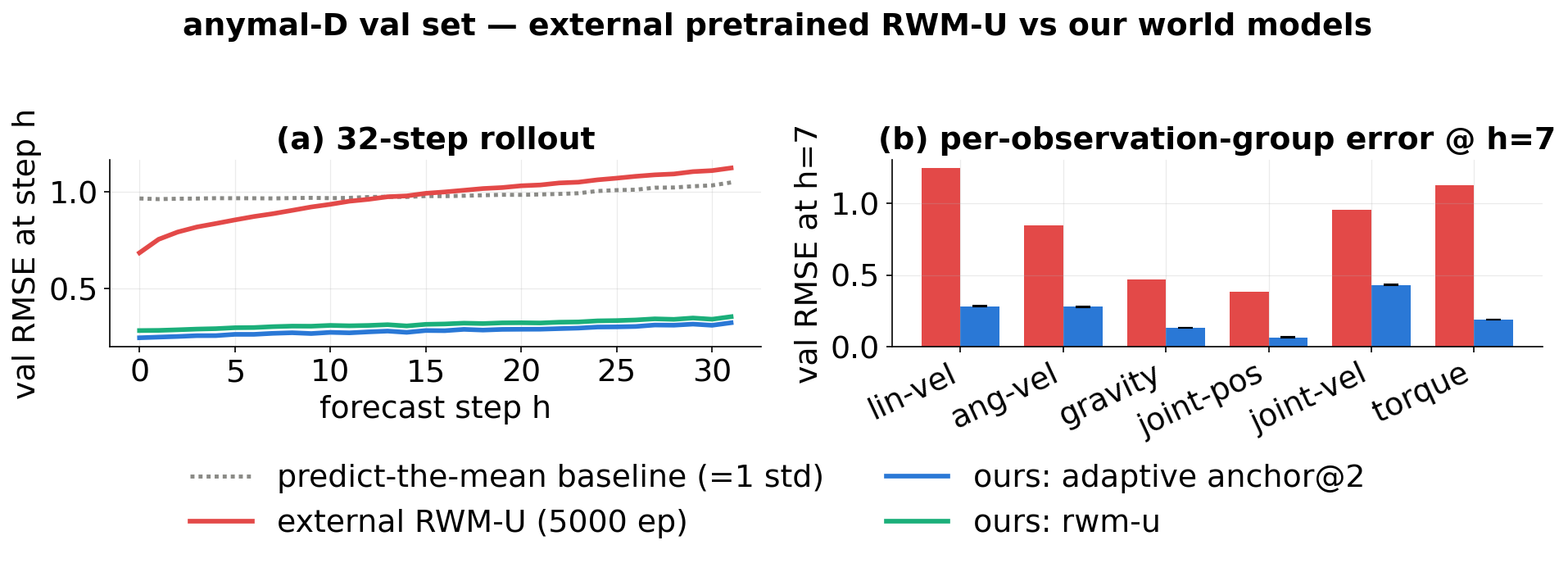}
    \caption{\textbf{Deploying the pretrained world model by \citet{li2025uncertainty} on our dataset.} We verified that the observed error is not an interface artifact. Under these conditions, the released checkpoint does not produce usable forecasts on our ANYmal-D distribution. Its error exceeds both of our models over the full 32-step rollout (a) and across every observation group at $h=7$ (b), and tracks the predict-the-mean floor throughout the horizon.}
    \label{fig:external_wm}
\end{figure}

The repository by \citet{li2025uncertainty} is available at
\texttt{leggedrobotics/robotic\_world\_model\_lite}.\footnote{\url{https://github.com/leggedrobotics/robotic_world_model_lite}} It provides a pretrained world model checkpoint. We validated the RMSE of this model with our results achieved on our dataset. However the provided model underperforms compared to our results, as shown in Fig. \ref{fig:external_wm}. The RMSE over the forecast horizon increases, matching the behavior of our results. As a sanity floor we plot the error of a constant predictor that always outputs the dataset mean. In our z-scored units this trivially attains an RMSE of one standard deviation ($\approx 1.0$), so it marks the boundary between predictive and non-predictive behavior. The externally pretrained RWM-U \cite{li2025uncertainty} tracks this line across the horizon, confirming that its forecasts transfer no usable dynamics to our ANYmal-D distribution, while our models remain lower.

\section{Compute savings}
\label{app:flops}
Table \ref{tab:compute} shows the compute measure in FLOPs for the different training schemes. Both the fixed-8 and the adaptive ensemble reduce the compute as already measured in cumulative rollout steps. But here, the adaptive method is the most efficient alternative. The reason fixed-8 has more FLOPs is that it still has to process the 32-history steps in each iteration. Since it needs more training iterations to reach the same amount of cumulative rollout steps, fixed-8 always operates in the inefficient short-horizon regime. It repeatedly pays the full history cost for only eight forecast steps.

\begin{table}[h]
    \centering
    \begin{tabular}{lcc}
    \hline
    Method & FLOPs [PFLOP] & $\Delta$ FLOPs \\
    \hline
    fixed-32          & 60.2 & --- \\
    fixed-8           & 21.4 & $-65\%$ \\
    RWM-U~\cite{li2025uncertainty} & 90.6 & $+51\%$ \\
    Adaptive \texttt{anchor@2} & 18.0 & $\mathbf{-70\%}$ \\
    \hline
    \end{tabular}
    \caption{\textbf{Training compute on ANYmal-D.} Fixed-32 and RWM-U report FLOPs at
    their own convergence, fixed-8 and adaptive \texttt{anchor@2} report the FLOPs to reach
    fixed-32's accuracy. FLOPs are measured GRU-inclusive.}
    \label{tab:compute}
  \end{table}

\section{Robustness}
\begin{figure}[h]
    \centering
    \includegraphics[width=\columnwidth]{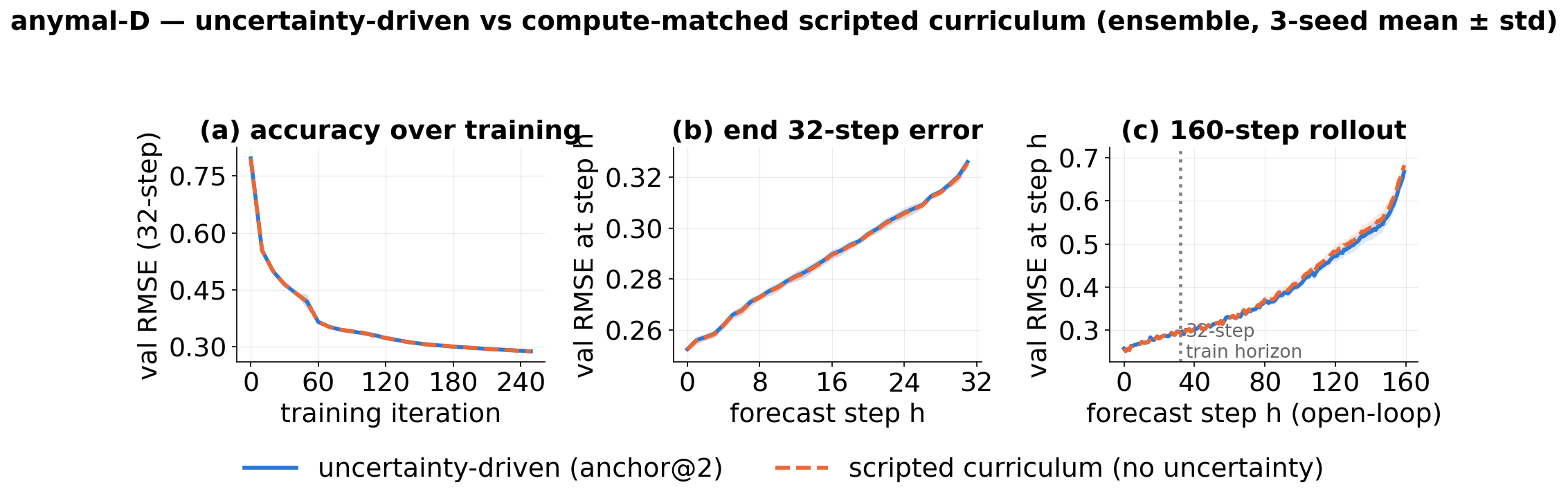}
    \caption{\textbf{Compute-matched non-uncertainty ablation.} Uncertainty-driven
  adaptive (\texttt{anchor@2}) vs.\ a scripted curriculum that replays each seed's
  recorded per-iteration rollout-length schedule with no uncertainty gating,
  matched in compute and identical in every other respect. (a) validation RMSE
  over training, (b) end-of-training per-step error over the 32-step horizon,
  (c) 160-step open-loop rollout. The two agree to within seed noise at every
  horizon: the accuracy gain comes from the induced curriculum, which epistemic
  uncertainty discovers automatically per run, with no manual schedule or oracle.}
    \label{fig:scripted}
\end{figure}

The threshold is a self-calibrating heuristic. The frozen anchor is not a coverage-calibrated threshold. It is an empirically-selected stopping rule, a robust (median) estimate of ensemble disagreement at a shallow reference horizon. Its purpose is to shape the training curriculum (which rollout depths receive gradient), not to certify deployment-time rollouts. Reliability is assessed empirically via long-horizon rollout error and policy sim-transfer. A coverage guarantee could be obtained by replacing the anchor with a conformal quantile calibrated on held-out rollouts. We leave this as an extension. Its lack of a formal guarantee is not a practical liability: an ablation replaying the induced horizon schedule without any uncertainty gating reproduces the adaptive model's accuracy to within seed noise across all horizons (Fig.~\ref{fig:scripted}), showing the method's benefit comes from the reproducible curriculum it induces. This does not diminish the role of uncertainty. The manually replayed schedule is an oracle derived from an adaptive run (\texttt{anchor@2}), whereas the baseline schedules without it (fixed-8, fixed-32, RWM-U) are strictly worse at short or long horizons (Fig.~\ref{fig:fixed32_anchor2}). Uncertainty is the mechanism that produces the effective schedule without tuning.

\end{document}